# A Brief Wellbeing Training Session Delivered by a Humanoid Social Robot: A Pilot Randomized Controlled Trial

Nicole L. Robinson[ab]*, Jennifer Connolly[c], Gavin Suddrey[a], David J. Kavanagh[c]

[a]Queensland University of Technology, Australian Research Council Centre of Excellence for Robotic Vision, Brisbane, Australia

[b]Monash University, Department of Electrical and Computer Systems Engineering, Turner Institute for Brain and Mental Health, Victoria, Australia

[c]Queensland University of Technology, Centre for Children's Health Research, Institute of Health and Biomedical Innovation, and School of Psychology & Counselling, Brisbane, Australia

**Abstract**

Mental health and psychological distress are rising in adults, showing the importance of wellbeing promotion, support, and technique practice that is effective and accessible. Interactive social robots have been tested to deliver health programs but have not been explored to deliver wellbeing technique training in detail. A pilot randomised controlled trial was conducted to explore the feasibility of an autonomous humanoid social robot to deliver a brief mindful breathing technique to promote information around wellbeing. It contained two conditions: brief technique training ('Technique') and control designed to represent a simple wait-list activity to represent a relationship-building discussion ('Simple Rapport'). This trial also explored willingness to discuss health-related topics with a robot. Recruitment uptake rate through convenience sampling was high (53%). A total of 230 participants took part (mean age = 29 years) with 71% being higher education students. There were moderate ratings of technique enjoyment, perceived usefulness, and likelihood to repeat the technique again. Interaction effects were found across measures with scores varying across gender and distress levels. Males with high distress and females with low distress who received the simple rapport activity reported greater comfort to discuss non-health topics than males with low distress and females with high distress. This trial marks a notable step towards the design and deployment of an autonomous wellbeing intervention to investigate the impact of a brief robot-delivered mindfulness training program for a sub-clinical population.

*Keywords:* Mindfulness Training, Social Robot, University Cohort, Young Adults, Human-Robot Interaction

# 1. Introduction

## 1.1 Global Mental Health in the Adult Population

One in four people in the world will experience a mental health condition, making mental disorders the leading cause of ill-health and disability worldwide [1, 2]. Mental disorder rates have risen across all age spans, including disability-adjusted life years [1]. For instance, 1 in 5 US adults experience mental illness [3]. However, two-thirds of people with a mental illness will not receive treatment [2], which contributes to significant long-term psychosocial problems, such as social isolation, job instability, and associated physical illness [1, 4-6]. There is an urgent need to scale up treatment for mental health services to deliver mental care to the global population that needs it. Current services do not meet treatment demands with a world median of 1.20 psychiatrists and 0.60 psychologists per 100,000 people [7, 8].

## 1.2 University Students and Higher Education Settings for Mental Health Rates

There has also been an increase in concern for mental health and wellbeing of students enrolled in college and university-level courses [9, 10]. The typical age period to start higher education (i.e. young adulthood) is also characterized by higher rates of serious psychological distress (71%) and major depressive episodes (13.2%) [11, 12]. In comparison to the general population, higher education students have higher average scores related to depression, anxiety and stress [13-16]. In addition, students who experience elevated mental distress and later screen positive for a mental disorder are at higher risk of suicidal behavior [17]. Students in higher education often do not seek or access mental health support services that are available to them. For example, less than one quarter of students reported they would seek treatment if they experienced psychological distress [18]. Counselling attendance can be considerably low (e.g. 10%), and three quarters of students who did report clinically significant distress levels did not receive counselling in the last six months [19]. Identified barriers for help-seeking for mental health included perceived stigma in relation to their level of emotional distress [20], disclosure about their mental state [21, 22], belief that stress is a 'normal' part of higher education, desire to deal with issues on their own, time to receive treatment [23] and the lack of knowledge around available services [21].

There is a need to both increase mental service offering and reduce identified barriers for accessing services. Higher education settings represent an important opportunity to support students with mental health and wellbeing challenges and provide an avenue for early intervention [24]. The provision of accessible support services in higher education is vital to assist students to manage the transition to university life, build resilience and develop coping strategies to manage stressors related to the pursuit of higher education, consequently reducing their future risk of a mental health disorder.

## 1.3 Digital Interventions

Digital mental health is an intervention modality designed to assist in reaching at-risk populations and to overcome barriers in accessing traditional support services [e.g. 25, 26, 27]. Mental health programs delivered through digital modalities have the potential to reach large populations [28] and to help create a gateway to encourage further help-seeking behaviours [29]. Digital interventions in mental healthcare have shown clinical benefits for anxiety and depression for young adults and adolescents [25, 28], as well as for specialist domains such as within the workplace [30].

The use of social robots to deliver healthcare tasks such as information, assessment, and intervention is gaining momentum as a novel method to engage people in treatment [e.g. 31]. Robots have interpersonal strengths compared to other types of digital-based methodologies, such as the ability to create a dialogue between the human and robot to create a healthcare plan [32, 33], increased perceived empathy between a robot and a person [34], lower perceived stigma and sense of judgement from a robot [35], and perceived responsiveness to personal disclosure increases willingness to use it during stressful events [36]. Robots emulate similar strengths compared to other digital-based interventions, such as reduced ongoing cost, adherence to treatment protocol, and ease of access (if the service is readily available) [37-39]. Initial work has been conducted to explore feasibility for robot use in components that would benefit components of psychotherapy, such as exploring the prevalence of self-disclosure to humanoid robots [40-42] and helping to reduce anticipatory anxiety and tension for interesting with a robot instead of a human [43]. There was also an increasing trend towards sharing more information and disclosures across a longer-term interaction period of 5 weeks [44].

Robots have been investigated in mental health support and treatment with positive evaluations when they are used as a tool in psychotherapy in adult and children samples [45-47]. Trials have been conducted to explore the efficacy of the intervention in health-related outcomes. A mixed-methods design was used to explore the impact of a socially assistive robot for low-income, socially isolated older adults and reported improved health-related quality of life and reduced depressive symptoms [48]. A pilot randomized controlled trial for a robotic assistant for children with cancer found reductions in stress, depression and anger scores [49]. Robots have been deployed to help mitigate stress, pain and anxiety for pediatric patients [50], and a robot-led distraction program reduced pain and distress in children who underwent a vaccination [51], including reduced distress for a subcutaneous port needle insertion [52]. Research trials have used animal-like robots to mimic the effects of animal assisted therapy to also contribute to creating health-related outcomes [53]. Such trials have yielded increased mood scores for patients with dementia after a 15-minute interaction [54], and improvement on apathy and irritability scores over 3 months for nursing home patients [55]. This included observed increases in positive affect and behavioral indicators alongside decreases in negative affect and behavioral indicators for veteran residents in a geropsychiatric long-term care facility [56].

## 1.4 Mindfulness-based Interventions

An intervention that has shown to be both translatable to digital delivery and beneficial for clinical and non-clinical samples is mindfulness-based treatment [57, 58]. Mindfulness has been found to improve a range of psychological issues for those who practice it, such as emotional reactivity, behavioral regulation, and subjective wellbeing [59]. Correlational research has found that mindfulness levels are positively associated with psychological health indicators, such as positive affect, life satisfaction and emotion regulation [59]. Positive effects for the reduction of symptoms for stress and anxiety can also be found from the use of a single mindfulness-related technique, such as the body scan [60]. Interventions that deliver mindfulness-based treatment content in single sessions under 30 minutes could create changes on a health-related outcome such as craving reduction, relaxation levels and reduced negative affect [61]. In a higher education student sample, mindfulness-based stress reduction has been associated with lower levels of mental distress and improved subjective wellbeing compared to a control group for those who scored high on neuroticism [62]. Digital delivery of mindfulness meditation for college students has been found to be effective at improving ratings on depressive symptoms, resilience, and college adjustment across a 10-day period [63].

Social robots have not often been explored as a method for mindfulness-based techniques, but there is evidence supporting mindfulness delivery using computer-based agents. Computer-based agents that have no physical embodiment (i.e. conversational agents) have been deployed in mental healthcare to support people to make wellbeing improvements. Text-based conversational agents have shown promising suitability and effectiveness for mental health applications, such as reduced psychological distress scores post-intervention [64]. This includes improvements on wellbeing and perceived stress scores for those who adhered to the intervention compared to a control group [65]. Embodied agents such as virtual coaches have also delivered common wellbeing techniques such as mindfulness meditation, and were found to be more effective in eliciting routine practice of the technique when compared to audio or written materials [66]. An automated program to teach mindfulness for wellbeing showed an impact on practice from pre- to post-intervention which was sustained at 3-months compared to control conditions [67]. Computer-based agents were also effective at teaching a broad range of lifestyle strategies (mindfulness, stress management, healthy eating, and physical activity) over a 1-month follow-up compared to patient information sheets [68]. These trials demonstrate that digital mindfulness-based interventions can create improvements across wellbeing dimensions, even when delivered in brief interventions, although this delivery method must also be paired with a modality that can reduce help-seeking barriers. There has also been some initial work for the use of a tele-operated robot to conduct mindfulness training, including to explore people's perception of a robot coach [69]. A robot-guided mindfulness practice assessed through EEG changes during a practice session also found that a robot coach could help people to achieve a mindful state [70, 71].

To summarize, significant levels of stress, anxiety and depression are present within the general population and within higher education settings. It is known that young people who experience emerging symptoms of stress, anxiety and depression are unlikely to seek traditional forms of support, suggesting a need for innovative methods to reach this vulnerable group. Digital health programs have demonstrated efficacy in supporting adults with their mental health and wellbeing in the absence of intervention from human clinicians. Other innovative technologies such as social robots are following a similar pathway and offer additional benefits over standard digital and computerized programs. Robots have been trialed as a digital tool to support mental wellbeing in some domains, and their use has been characterized as helpful when teaching people to use health-related techniques. Social robots in health services are frequently rated as entertaining, engaging and personable, and therefore can conceivably be programmed and deployed to provide a level of support within higher education settings. A robot readily available on a university campus presents an opportunity to teach students a brief wellbeing technique that they can use in their own time to manage stress or mild anxiety. The availability of the robot and sense of reduced stigma could overcome some of the barriers to engagement with traditional support services. A pilot randomized controlled trial was designed as a starting step to explore engagement with, as well as acceptability and perceived usefulness of a robot to support mental wellbeing for those in a higher-education setting, prior to deployment of a larger trial.

## 1.5 Trial Design

This was a pilot randomized controlled trial to investigate the utility and acceptability of an autonomous humanoid social robot to deliver a brief mindful breathing meditation. The purpose of this trial was to explore the feasibility of the future development of a longer-term robot-delivered wellbeing program, and to develop into a larger longitudinal trial. This included exploring feasibility steps as listed by exploring dimensions such as acceptability, demand, implementation, practicality, integration, and limited efficacy [72]. The trial had two experimental conditions: brief mindfulness technique training ('Technique') and conversational control designed to represent a simple wait-list activity which involved the robot asking closed-ended questions to represent building up a communicative relationship between the robot and the person ('Simple Rapport'). The trial was implemented to provide insight into three key research questions:

- **Primary:** Explore intervention effects on mood, incentives to use a robot, intention to continue to use a social robot in a healthcare context, and comfort and likelihood to discuss topics with a social robot including health-related information,
- **Secondary:** Examine effects of gender and distress on response to being trained in a wellbeing technique delivered by a social robot and,
- **Secondary:** Assess the receptiveness to a robot-delivered program on a university campus and viability of recruitment

It was predicted that both conditions would produce high ratings on the robot evaluation scales, indicating acceptability to receive wellbeing technique training from a social robot compared to a simple conversational wait-list activity. It was predicted that acceptable recruitment rates would establish feasibility to run a larger trial in future. Human research ethical approval was obtained, and trial recruitment occurred over 6 months.

## 2. Methods

### 2.1 Target Participant Group

Prospective participants were recruited at a higher education campus and included university staff and students as well as members of the general public who were visiting the campus. Prospective participants needed to be aged 18 years or older and consent to attending a 10-minute session in a private room. No affiliation with the university was required for participation. Recruitment methods included convenience sampling, noticeboard flyers, word-of-mouth, and social media posts.

### 2.2 Participant Sample

A total of 241 participants provided consent and started the session. Two (1%) withdrew, leaving 239 participants to complete the session (1 found the interaction difficult; 1 declined to continue in the second half of the session). Nine outliers were removed based on z-scores above 3 on robot evaluation scores, leaving a total of 230 participants with complete data. Condition was randomized somewhat evenly between 'Technique' (n = 106, 46%) and 'Control' (n = 124, 54%). A minor tablet-related error occurred with the interface, but given its minimal impact on trial outcome and the individual's interest to finish the session, the data point was not deemed necessary to remove. A total of 221 participants (92%) from the full sample obtained some form of compensation for their time: 186 (78%) took part in the prize draw and 35 (15%) received course credit. From a subsample of 347 prospective participants informed about the trial, 179 declined and 4 were underage, yielding on average a 53% uptake rate.

*2.3 Descriptives*

There was a relatively even split of female (n = 108, 47%) and male (n = 122, 53%) participants with a mean age of 29 years ($SD$ = 11.77, Range = 18–67). They were mostly single (n = 107, 46%) or in a relationship (n = 60, 26%). A large percentage were university students (n = 164, 71%). Many had completed higher education (n = 128, 56%), such as a certificate (n = 26, 11%), trade (n = 3, 1%), undergraduate (n = 58, 25%) or post-graduate degrees (n = 57, 25%). Most were currently employed (n = 167, 73%) either in full-time (n = 53, 32%), part-time (n = 47, 28%) or casual work (n = 67, 40%). There was a low reported level of experience with programming ($M$ = 3.23, Mode = 0, $SD$ = 2.94, Range = 0-10) and robotics ($M$ = 2.00, Mode = 0, $SD$ = 2.44, Range = 0-10). The K-10 mean score was 21.58 ($SD$ = 6.25) with 36 (16%) in the low category, 90 (39%) in moderate, 75 (33%) in high and 29 (13%) in very high. Participants completed sessions in 10 minutes on average ('Technique' $M$ = 10.70, $SD$ = 2.30, 'Control' $M$ = 10.72, $SD$ = 1.75). There were no significant differences between conditions on any demographic variables.

*2.4 Robot System Architecture*

A Pepper Humanoid Robot by SoftBank Robotics delivered the trial [73]. Pepper is 1.21 meters tall and weighs 28 kilograms. It has two 5-megapixel cameras, two speakers, and five tactile sensors. Pepper has an LG tablet (24cm x 17cm x 14.5mm) connected to its chest and an overall battery life that can last several hours without charge, allowing for continuous testing sessions. Pepper was programmed via the NAOqi 2.5.5 Operating System using a custom-built HTML/JavaScript service [74]. The robot was equipped with several packages from the NAOqi 2.5.5 library: ALAnimatedSpeech, ALTextToSpeech, ALFaceTracker, and ALAutonomousLife without modifications built into the libraries. The robot implemented a rule-based system to deliver the interaction and to collect trial data without the need for human involvement (i.e. no Wizard of Oz operation). The interaction involved scripted segments which included both short verbal monologues paired with gestural animations. The robot used co-verbal gesturing, and each gesture was chosen to best reflect the intended message and instruction that was being presented at the time [75]. The robot spoke each question out loud before displaying the associated text to help mimic a more natural conversation style and direct attention away from proactively fixating on its screen. Participants responded to the questionnaire sets through the tablet using input elements such as radio buttons, check boxes and text box entries. Participants controlled the interaction using navigation buttons to transition to the next segment once they had finished their response. Speech recognition and language processing was not used for data collection because it can increase the likelihood of inaccurate data capture, and data entry through a tablet interface represents a more robust method for long-term deployment in a wellbeing intervention. Participants were asked to complete the short session without robot training, and they were not expected to have any prior experience with voice commands.

## 2.5 Technique Condition

The robot provided a brief monologue about the importance of wellbeing and asked for permission from the person to talk further about it. If agreed, the robot provided information about the use of brief wellbeing techniques and how a short mindfulness technique can have some benefits if it is practiced regularly over a period of time. The robot asked permission to teach the participant a brief mindfulness-based exercise, and if agreed, participants took part in a 1-minute guided practice focused on mindful breathing, followed by 1 minute to practice on their own while the robot displayed a timer wheel to assist in the count down. The description of mindfulness and the brief guided practice could both be skipped if the individual declined when the robot sought permission.

## 2.6 Control Condition

This condition was designed to create a time-matched control against the wellbeing content and measures. The robot provided information about itself, including details about its name, height, weight, features, and role in the research centre. The robot provided information about its prospective use in healthcare, including collecting patient data, teaching brief wellbeing techniques, and the provision of health-related advice. After this, the robot asked individuals three questions about themselves (How did you get here today, which animal would you like to have as a pet, and what season do you like the best). The robot gave a closed answer set for them to choose their response (i.e. car, bike, bus, walk, train, ferry). To conclude, the robot provided a basic reworded summary of their responses back to the person out loud at the end of their questionnaire set, and participants were asked to confirm if the robot correctly summarized their answers [Yes/No]. This interaction was designed to both be a waitlist control activity against the technique training condition, but also to represent a simple relationship-building session to become more familiar with the robot and its ability to interact with a person.

## 2.7 Experiment Reliability and Validity

Simple randomization for condition allocation was blinded from the research assistant through the use of a hidden electronic function at the start of each session. Simple randomization was considered to be an acceptable method given the intended sample size was above 200 people [76, 77]. Randomization simulations were run prior to deployment which found the function would approximate ~50/50 condition allocation. Completed responses were uploaded as a stream of JSON data to a service running on the robot which removed possible identifiers from the collected data (i.e. removal of time stamps). All responses were saved to a secure password protected file. Trial data was retrieved from the robot via a secure shell session, moved to a secure storage location, and then deleted from the robot after each data extraction session. Electronic access to the experimental program was protected and restricted to the research team only. Physical access to the robot was controlled to ensure the integrity of the trial, including storing the robot in a locked storage room when not in use. Prior to deployment, the interaction was tested by the research team, roboticists, and volunteers not associated with the project across a minimum of 15 test runs. Trial data during these test sessions were reported manually on non-digital methods and compared to digital counterparts stored in the JSON file. This included testing different options and possible response combinations across each trial run. This method showed that there were no translation errors from initial data input through to the final digital data file, and all trials were recorded and stored correctly during the testing process. Once deployed, no hardware or software modifications were made, and all participants received the same application script of this trial.

## 3. Measures

### 3.1.1 *Demographics and Technique Condition Questionnaire*

Demographic data included age, gender, relationship status, highest completed level of education, if they were currently studying and their study area, employment status, and area of employment. Participants were asked to report their level of experience with programming and robotics on an 11-point scale (0 = No experience at all, 10 = Highly experienced). Participants were asked to complete a brief set of questions about their state of mood (relaxed, content, focused) before and after condition content (0 = Not at all, 10 = Extremely), which represented a simple and brief version of an affect change scale. Participants were asked to rate the brief guided mindfulness meditation for: level of enjoyment; usefulness; and likelihood to use it again (0 = Not at all, 10 = Extremely). Participants were asked if they had previous experience with mindfulness and if they responded yes, they were asked how often did they practice: daily, weekly, fortnightly, monthly, every few months or only once or twice.

### *3.1.2 Kessler Psychological Distress Scale*

The Kessler Psychological Distress Scale (K-10) [78] is a brief self-report measure to identify the likelihood of a psychiatric disorder and the need to seek support from a mental health service [78]. The scale has 10 items measuring distress in the past 4 weeks across four dimensions: anxiety, tiredness, agitation and depression. Each item is rated on a 5-point scale (1 = None of the time, 5 = All of the time) and added together to provide a total score (10 to 50 points). The scale was scored using the Andrews and Slade [79]'s scoring format: 10-15 as low, 16-21 as moderate, 22-29 as high, and 30-50 as very high. The K-10 demonstrates good to excellent internal consistency in diverse populations (Cronbach's alpha > 0.88) [e.g. 80]. For current study K-10 scores were split into low (≤21) vs high (≥22).

### *3.2 Robot Evaluation Scales*

The Robot Incentives Scale (RIS) [81] measures perceived incentives to engage with a social robot. It includes three subscales: 'Emotion' with 5 items for its likability, 'Social' with 3 items for social/relational aspects, and 'Utility' with 4 items for perceived utility. Each item is rated on an 11-point scale (0 = Not at all, 10 = Definitely). Cronbach's alphas for each subscale in the current study were good to excellent (Emotion = .93; Social = .88; Utility = .92).

The Robot Usage Intention (RUI) is a 5-item question set (0 = Not at all, 10 = Definitely) assessing how willing people would be to interact with the robot. The scales have been tested on different age range samples (adolescents to older adults), and shown to be sensitive to change across multiple timepoints The scale can assist in the prediction for willingness to engage social robots both in the short and long-term [81]. Cronbach's alpha in the current study was excellent ($\alpha$ = .95).

The Robot Disclosure Questionnaire was a custom-built set of 10 items designed to measure how likely (Likely, 5-items) and comfortable (Comfort, 5-items) an individual would feel to talk to a social robot about different topics including: 1) casual conversation topic, 2) solving a problem or help with a task, 3) getting advice/support on a sensitive topic, 4) medical symptoms or conditions, and 5) mental health symptoms or conditions. All items were measured on an 11-point scale (0 = Not at all, 10 = Definitely). It has previously been tested in a prior human-robot interaction trial [81-83]. In the current study, the items for both Likely and Comfort were summed into two subscales representing likelihood or comfort to discuss health (medical symptoms or conditions and mental health symptoms or conditions) and non-health (casual conversation topics, solving a problem or help with a task, and getting advice/support on a sensitive topic) topics. Cronbach's alpha for the 3-item non-health scales were acceptable (Likely Non-Health = .80; Comfort Non-Health = .74). The two-item health scales yielded high Spearman Brown coefficients (Likely Health = .84; Comfort Health - .86).

### 3.3 Procedure

Prospective participants were recruited by a research assistant who gave a detailed information sheet and brief outline of the proposed trial. Participants provided digital consent on the robot tablet interface, and randomization occurred before the start of each session. Participants completed the interaction with the robot alone to minimize the potential of researcher effects or biased responding. The researcher waited outside the room in the event of technical difficulties or participants wishing to seek clarification. Tablet questionnaires were presented to collect trial measures and limited to minimize burden, keeping the total participation time under 10 minutes. After completing the experiment, participants were given the option to provide an email to enter a prize draw or an identification code to receive course credit at the end of each session. If individuals chose to receive course credit or entry to a prize draw, this data was stored in separate files to protect response identification. Entry into the draw or receipt of course credit was voluntary and optional, so participants could complete the session without providing an email address or identification code. Once participants left the testing room, they were thanked for their time and asked if they had any questions about their participation. Any events disclosed to the research assistant were recorded, such as reported technical faults. A photograph of the experimental set-up can be seen in Figure 1.

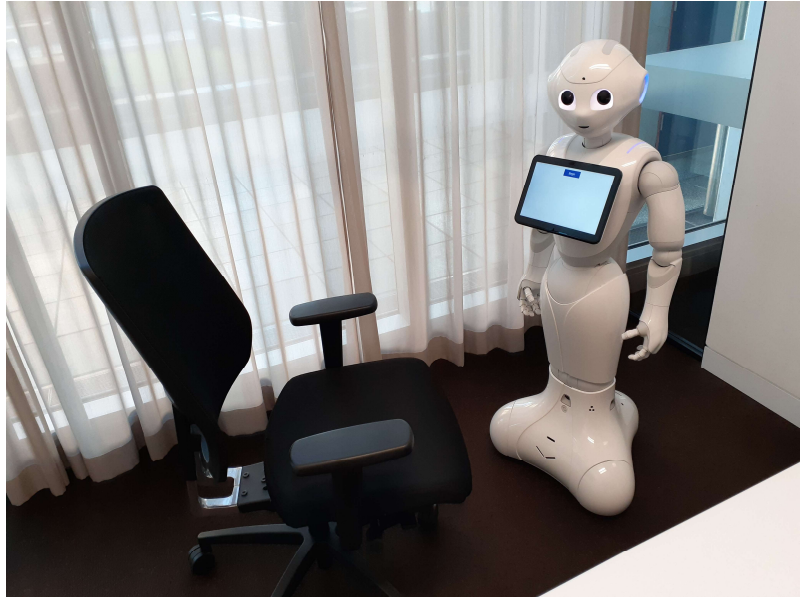

**Figure 1: Experimental Setup**

## 4. Results

### *4.1 Intervention Effects – Overall Descriptives*

In the 'Technique' condition, 103 individuals (97%) agreed to hear information about mindfulness and of those, 102 (99%) accepted to complete a brief exercise. There were moderate to high ratings of enjoyment of the technique ($M = 7.20/10$, $SD = 1.61$), of its perceived usefulness ($M = 7.35/10$, $SD = 1.85$), and likelihood to repeat using the mindfulness technique again ($M = 7.62/10$, $SD = 2.00$). In this condition, 72 participants

(71%) had tried mindfulness before. Of these, there was a relatively even split across frequency of use: daily (14%), weekly (18%) fortnightly (11%), monthly (15%), every few months (19.4%) and only once or twice (22%). The 30 participants who had never done a brief mindfulness exercise before reported moderate enjoyment ratings ($M = 6.97$, $SD = 1.38$), perceived usefulness ($M = 6.90$, $SD = 1.65$) and likelihood to repeat the exercise again ($M = 6.70$, $SD = 1.78$). In the 'Control' condition, 100% of participants (n = 124) confirmed that the robot correctly summarized their answers for their chosen mode of transport, favorite animal and season.

### 4.2 Intervention Effects – Conditions Only

There were no significant time x condition effects in ratings from pre to post interaction for feeling more relaxed, content, or focused. There were significant time effects across all variables with participants reporting a significant increase in ratings from pre to post interaction (content: $F(1, 228) = 55.30$, $p < .001$; relaxed: $F(1,228) = 107.14$, $p < .001$; focused: $F(1,228) = 18.97$, $p < .001$).

### 4.3 Pre to Post Interaction Scores for Condition Evaluation with Effects from Condition and Gender

There were significant time effects across all mood variables, with participants reporting a significant increase in ratings from pre to post interaction (content: $F(1, 228) = 58.34$, $p < .001$; relaxed: $F(1,228) = 101.68$, $p < .001$; focused: $F(1,228) = 19.62$, $p < .001$). For contentment, there was also a significant time x condition x K-10 effect ($F(1,228) = 4.29$, $p = .039$), with participants who had low K-10 scores and received the mindfulness intervention increasing their contentment ratings more than people who had low K-10 and received rapport. A significant effect of K-10 score for contentment ratings, was also found ($F(1,228) = 9.80$, $p = .002$) with participants with low K-10 scores rating higher levels of contentment across both time points.

For ratings of relaxed mood, there were significant main effects for gender ($F(1,228) = 4.02$, $p = .046$) and K-10 ($F(1,228) = 11.51$, $p < .001$). Males reported higher relaxation ratings at both pre and post as did participants with low K-10 scores. A significant interaction between gender and K-10 score was also found ($F(1,228) = 4.78$, $p = .030$) with females with high K-10 scores reporting lower relaxation scores relative to males and to females with low K-10 scores. No significant effects were observed for focus ratings beyond the time effect reported above.

### 4.4 Mood ratings (K10 Scores) with Effects from Condition and Gender

Repeated measures ANOVAs were used to examine the effects of condition, gender and K-10 score on mood ratings pre and post interaction. Mean ratings by group are presented in Table 1.

Table 1. Pre and post interaction mood ratings by gender and K-10 score between conditions.

|  |  | Control | | | | | | Mindfulness | | | | | |
|---|---|---|---|---|---|---|---|---|---|---|---|---|---|
|  |  | Low K-10 | | | High K-10 | | | Low K-10 | | | High K-10 | | |
|  |  | N | M | SD | N | M | SD | N | M | SD | N | M | SD |
| Content | M pre | 39 | 7.72 | 1.52 | 20 | 6.40 | 2.89 | 35 | 7.37 | 1.77 | 28 | 7.21 | 1.81 |
|  | M post | 39 | 7.97 | 1.86 | 20 | 7.50 | 2.48 | 35 | 8.11 | 1.23 | 28 | 7.75 | 1.38 |
|  | F pre | 30 | 7.83 | 1.98 | 35 | 7.17 | 1.52 | 22 | 7.32 | 2.08 | 21 | 6.52 | 1.54 |
|  | F post | 30 | 8.47 | 1.53 | 35 | 7.86 | 1.56 | 22 | 8.27 | 1.42 | 21 | 7.10 | 1.79 |
| Relaxed | M pre | 39 | 7.56 | 1.57 | 20 | 7.15 | 2.18 | 35 | 7.46 | 1.84 | 28 | 7.14 | 1.35 |
|  | M post | 39 | 8.03 | 1.71 | 20 | 7.90 | 1.83 | 35 | 8.49 | 1.22 | 28 | 8.29 | 1.21 |
|  | F pre | 30 | 7.67 | 2.09 | 35 | 6.43 | 2.13 | 22 | 7.32 | 2.15 | 21 | 5.90 | 1.48 |
|  | F post | 30 | 8.43 | 1.63 | 35 | 7.49 | 1.81 | 22 | 8.27 | 1.72 | 21 | 7.00 | 1.70 |
| Focused | M pre | 39 | 7.56 | 1.55 | 20 | 7.15 | 1.57 | 35 | 7.26 | 1.95 | 28 | 7.29 | 1.92 |
|  | M post | 39 | 7.51 | 1.75 | 20 | 7.55 | 1.36 | 35 | 7.71 | 1.45 | 28 | 7.82 | 1.31 |
|  | F pre | 30 | 7.37 | 2.37 | 35 | 7.06 | 2.21 | 22 | 7.50 | 1.44 | 21 | 6.24 | 2.30 |
|  | F post | 30 | 8.13 | 1.72 | 35 | 7.54 | 2.11 | 22 | 8.00 | 1.41 | 21 | 6.90 | 2.66 |

## 4.5 Robot Evaluation Scales with Effects from Condition and Gender

Two-way MANOVAs were used to examine effects of condition, gender and K-10 scores on the Robot Emotion, Social and Utility subscales. There were significant main effects of both condition and K-10 on Robot Utility scores ($F(1, 228) = 7.29$, $p = .007$ and $F(1,228) = 5.92$, $p = .016$ respectively). Participants in the control condition reported higher utility scores as did participants with high K-10 scores. There was a significant gender x K-10 interaction for the Robot Emotion subscale ($F(1,288 = 5.34$, $p = .022$) with men with high K-10 scores rating Robot Emotion higher than men with low K-10 scores, while the opposite was observed for women. No significant effects were observed for the Robot Social subscale. Means for each subscale are presented in Table 2.

Table 2. Descriptive statistics by condition, gender and distress levels for subscales of Robot Incentives Scale.

|  |  | Control | | | | | | Mindfulness | | | | | |
|---|---|---|---|---|---|---|---|---|---|---|---|---|---|
|  |  | Low K-10 | | | High K-10 | | | Low K-10 | | | High K-10 | | |
|  |  | N | M | SD | N | M | SD | N | M | SD | N | M | SD |
| Emotion | Male | 39 | 37.23 | 10.63 | 20 | 39.35 | 7.25 | 35 | 36.94 | 10.62 | 28 | 40.75 | 7.31 |
| Emotion | Female | 30 | 43.80 | 6.19 | 35 | 38.46 | 9.80 | 22 | 37.59 | 11.78 | 21 | 36.95 | 10.35 |
| Social | Male | 39 | 16.95 | 8.64 | 20 | 20.65 | 5.76 | 35 | 18.26 | 6.46 | 28 | 18.46 | 8.50 |
| Social | Female | 30 | 20.20 | 7.54 | 35 | 17.89 | 6.48 | 22 | 17.50 | 7.20 | 21 | 18.86 | 6.41 |
| Utility | Male | 39 | 25.18 | 10.10 | 20 | 28.85 | 6.57 | 35 | 22.11 | 8.01 | 28 | 27.14 | 6.89 |
| Utility | Female | 30 | 27.87 | 9.13 | 35 | 25.86 | 8.34 | 22 | 20.73 | 9.49 | 21 | 25.29 | 7.93 |

Three-way ANOVAs were used to explore effects of condition, gender, K-10 score and their interactions on the remaining robot scales. The means and standard deviations for these scales are presented in Table 3. There were no significant main or interaction effects for condition, gender or K-10 scores for the Robot Usage Intention Scale.

Table 3. Descriptive statistics by condition, gender and distress level for Robot Usage Intention and comfort and likelihood to discuss health and non-health topics.

|  |  | Control | | | | | | Mindfulness | | | | | |
|---|---|---|---|---|---|---|---|---|---|---|---|---|---|
|  |  | Low K-10 | | | High K-10 | | | Low K-10 | | | High K-10 | | |
|  |  | N | M | SD | N | M | SD | N | M | SD | N | M | SD |
| Robot Usage Intention | Male | 39 | 30.21 | 12.78 | 20 | 34.20 | 7.51 | 35 | 27.91 | 9.94 | 28 | 34.29 | 8.50 |
|  | Female | 30 | 35.27 | 10.79 | 35 | 31.66 | 12.04 | 22 | 28.18 | 12.18 | 21 | 31.52 | 11.21 |
| Comfort Health | Male | 39 | 13.00 | 5.44 | 20 | 15.05 | 4.85 | 35 | 14.66 | 4.47 | 28 | 15.18 | 3.37 |
|  | Female | 30 | 14.30 | 4.74 | 35 | 13.71 | 4.61 | 22 | 12.86 | 5.54 | 21 | 13.24 | 4.59 |
| Comfort Non-Health | Male | 39 | 20.18 | 6.35 | 20 | 24.15 | 4.44 | 35 | 21.94 | 4.46 | 28 | 23.50 | 4.48 |
|  | Female | 30 | 23.47 | 5.43 | 35 | 20.51 | 6.42 | 22 | 19.82 | 6.79 | 21 | 21.19 | 5.33 |
| Likely Health | Male | 39 | 12.82 | 5.18 | 20 | 14.65 | 4.60 | 35 | 13.83 | 4.33 | 28 | 15.00 | 3.46 |
|  | Female | 30 | 13.77 | 5.35 | 35 | 13.31 | 5.10 | 22 | 12.64 | 5.78 | 21 | 12.86 | 4.82 |
| Likely Non-Health | Male | 39 | 20.41 | 6.61 | 20 | 23.3 | 3.95 | 35 | 21.00 | 5.03 | 28 | 22.61 | 5.04 |
|  | Female | 30 | 22.40 | 6.63 | 35 | 20.11 | 7.01 | 22 | 19.18 | 7.33 | 21 | 20.33 | 5.64 |

## 4.6 Comfort and Likelihood to Discuss Health Topics with a Robot with Effects from Condition and Gender

Across participants, mean scores for comfort to discuss health topics ($M = 13.98$, $SD = 4.75$) were significantly lower than scores to discuss non-health topics ($M = 21.83$, $SD = 5.68$; $t(229) = 28.55$, $p < .001$). Similarly, mean scores for likelihood to discuss health topics ($M = 13.58$, $SD = 4.86$) were significantly lower than scores to discuss non-health topics ($M = 21.11$, $SD = 6.12$; $t(229) = 26.83$, $p < .001$). In the 3-way ANOVA for comfort to discuss non-health topics, the three main effects were not significant however two of the interactions were. There was an interaction between gender and K-10 ($F(1,229) = 4.69$, $p = .031$) with men with high K-10 scores reporting greater comfort discussing non-health topics with a robot than women with high K-10 scores (see Figure 2).

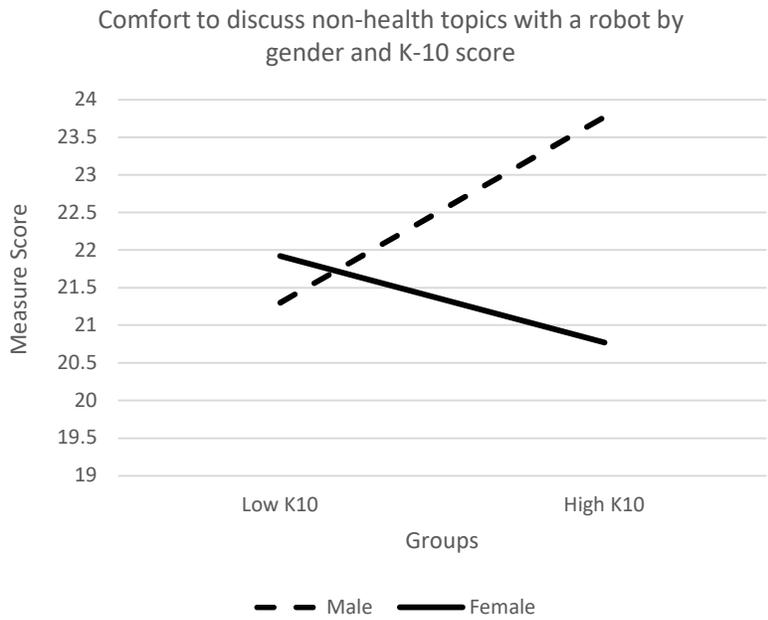

Figure 2. Comfort to discuss non-health topics with a robot by gender and K-10 score.

There was also a significant 3-way interaction between condition, gender and K-10 ($F(1,229) = 4.18$, $p = .042$) with associations between gender and K-10 differing across conditions (See Figure 3). Men who received the rapport condition and had low K-10 scores rated comfort lower than men with high K-10 scores who received rapport. Women demonstrated the opposite effect with women in the rapport group with low K-10 scores rating comfort higher than women in the rapport group with high K-10 scores. Men and women who received the mindfulness intervention showed similar patterns across K-10 scores.

Comfort to discuss non-health topics with a robot by condition, gender and K-10 score. There were no significant main or interaction effects for comfort to discuss health topics nor for likelihood to discuss health nor non-health topics.

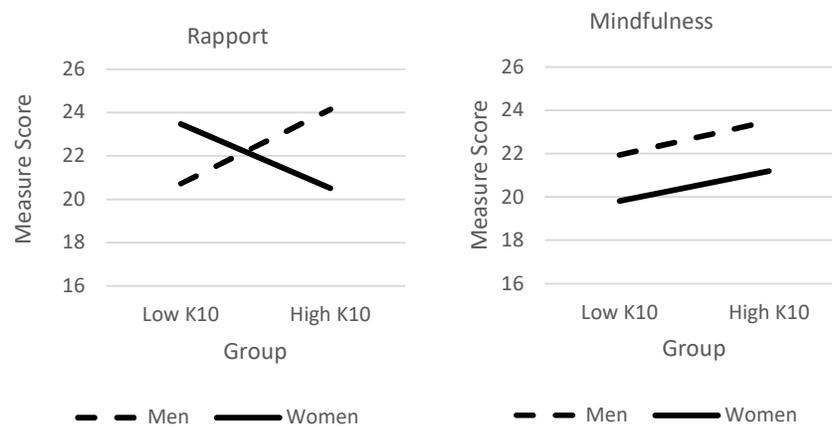

Figure 3. Comfort to discuss non-health topics with a robot by condition, gender and K-10 score.

## 5. Discussion

This trial explored the use of an autonomous humanoid robot to deliver a 10-minute interaction to facilitate wellbeing technique rehearsal in the form of a mindful breathing exercise, compared to a short interactive session designed to act as a relationship-building activity. Recruitment uptake and session completion rates were high, showing support for the uptake, acceptability and receptiveness around the use of a robot-delivered program on a university campus. Those who were willing and interested to complete a short mindfulness exercise lead by the robot rated the technique moderate to high in terms of enjoyment, perceived usefulness and likelihood to repeat the technique again, including for those who had never tried a mindfulness-based technique before. High reception, evaluation and proposed uptake of technique practice demonstrates that a social robot can be an effective way to teach people a brief wellbeing technique that people can practice in their own time. This includes helping to give higher education students access to an interactive modality to receive mental health information and learn new techniques that can help to improve wellbeing with practice.

Both conditions saw an increase in scores across timepoints for self-reported ratings of contentment, relaxation and focus during the session. There was a significant time x condition x K-10 interaction for contentment, with participants who reported low distress levels and who received the wellbeing technique reporting greater contentment post-exposure than participants with low distress who were allocated Control. This suggests a brief wellbeing technique may be more appropriate for people with low pre-existing levels of distress, while a longer intervention may be required for people experiencing elevated psychological distress. There was also an interaction effect between gender and distress levels on relaxation, with females with high distress reporting lower relaxation scores than other groups. Combined with the knowledge that depression is more prevalent in women than in men [84], this finding suggests that women may be a particularly important group to target with novel and innovative methods to prevent and reduce high distress levels.

There were no significant effects on the perception of robot sociability but there were for likability and utility. There was a significant interaction between gender and K-10 scores for the Emotion subscale with men with high K-10 scores reporting greater likability than men with low K-10 scores while the opposite was true for women. It is well documented that men access mental health treatment at rates much lower than women and hold negative views about help seeking [85, 86]. The fact that distressed men reported the highest robot likability ratings suggests that men may hold more positive attitudes to engaging with an embodied agent for support and assistance rather than human health professionals. Social robots may therefore offer a significant advantage in overcoming some of the barriers to male mental health support.

There were significant differences between conditions on robot utility ratings with participants who received Control rating utility higher than participants who received Technique. This suggests that answering questions about themselves and hearing a summary of their answers could have created a greater sense of perceived utility through seeing more of the robot's functionality. The control condition had an additional element of tablet interactivity by selecting answers compared to the technique session where individuals were asked to passively follow along with a static set of instructions. This additional layer may have provided greater insight to individuals about the robot's capability to customize based on the user's feedback, subsequently increasing scores related to perceived utility of the robot when later applied to a healthcare context, particularly if participants feel they are more involved in their healthcare plan. Participants with high K-10 scores also rated Utility higher than participants with low K-10 scores, regardless of condition. This may reflect a perception of potential for practical and/or emotional support.

Across all participants, ratings for comfort and likelihood to discuss health topics was significantly lower than those to discuss non-health topics. Despite high apparent acceptability and engagement, some people may remain reluctant to utilize semi-autonomous embodied agents for health-related concerns compared to more general conversational topics, especially after a relatively short interaction [87, 88].

In comfort to discuss non-health related topics, there was an interaction between condition, gender and K-10 scores, with comfort differing based on gender and distress levels only for people who received the Control condition. Men with high distress who received Control reported greater comfort to discuss non-health topics with the robot than men with low distress who received Control. The opposite was true for women, with low distress and receiving Control being associated with greater comfort than for women with high distress who received control. This seems to support earlier assertions about gender differences in help seeking attitudes and behaviours and suggests that while women may be an important focus for innovative interventions given their greater prevalence of depressive disorders, men may be more open and comfortable to work with alternative support modalities such as humanoid robots.

### 5.1 Design of a Robot-Delivered Wellbeing Session

Neither condition clearly outperformed the other, but equivalent scores showed that the potential use cases of a robot for wellbeing promotion or building a brief engagement session prior to a healthcare use case. Comparative condition results showed that individuals were as likely to rate the robot favorably in terms of perceived enjoyment, sociability, and likelihood to engage, irrespective of allocated condition. In other words, a brief wellbeing exercise was largely just as favorable as completing a brief conversation, and neither content had a significant differential impact on outcomes. This signifies that the evaluation of the robot itself may have had more importance on willingness to talk about health and non-health topics than the content itself, that personal factors may have played a more important role, or that it was their first robot exposure and therefore the content itself played a less salient role in evaluation. Initial impressions of robots might also have been made relatively quickly and within the first period of engagement. This outcome does demonstrate that the use of a social robot for wellbeing technique promotion and rehearsal is not any less acceptable than the general control condition. Given the importance of involving more higher education students and young adults in wellbeing promotion and mental health-related activities, a robot may present itself to be an advantageous method to achieve this.

Irrespective of condition content, time with the robot appeared to help increase states such as contentment, relaxation and focus over time. In this instance, this result may be more indicative of initial hesitation or indecisiveness on how to interact and interpret a social robot interaction decreasing over time, especially given the low group rating of robot experience in the sample. It is possible that the brief wellbeing technique was not powerful enough to create higher relaxation sensations given the short timeframe to practice if some hesitation on technology use was present. Increased scores may have been through a different route, such as improved familiarity with the technology over the brief timeframe. Therefore, adaptability and familiarity time might best be used as a method to allow time to adjust prior to delivery of more intensive content.

People were not more inclined to discuss health related topics in either condition, suggesting that spending time on a brief casual conversation with people prior to disclosure of health-related topics might not be essential. This is something to be explored further in future trials to substantiate this claim and to investigate methods that may increase willingness to discuss health topics, given the lower scores observed here for discussing health versus non-health topics. Alternatively, it is possible that additional variables caused noise in this evaluation. Those with higher social anxiety or hesitancy to disclose health-related symptoms may have preferred a robot that built some element of rapport first, whereas those who were time-poor or who wanted a method to encourage them to practice and feel accountable to finish the session may have preferred the initial techniques practice to complete. Exploration into the additional value of short interactivity with the robot prior to the interaction is warranted, and whether these lead to increased subjective or objective disclosure rates during the program.

## 5.2 *Implications for Mental Health and Wellbeing Support*

The automated training session for promotion of wellbeing advice has noteworthy implications for the creation and deployment of social robots to assist in wellbeing information and technique rehearsal. Mental health and wellbeing in young adults is an important target, given the high prevalence of psychosocial stress and incidence rates of psychiatric disorders [89], and problematic mental wellbeing for students within higher education [9, 13-16]. Participation in brief wellbeing practice led by a robot may help to reduce the initial entry barrier for those who are seeking some support, or who wish to later attend other support services. For instance, fear of disclosure about their mental health status or that they do not have time to complete treatment are commonly reported [21, 23], but uptake and completion rates in the robot-delivered intervention did not experience similar patterns. This could include referral to a longer session with a higher education counsellor, translation to a digital method to later continue practice, or longer-term wellbeing program provided by a clinician. However, it should be noted that similar difficulties might be seen if the robot intervention was positioned as a treatment for mental health disorders, and the robot program will interface with other programs, such as the healthcare clinic. Given that there are some issues around adherence and completion rates for internet-based interventions for students [90], a robot-delivered intervention may encounter similar long-term issues, but initial uptake rates were strong, representing at least a lower entry barrier to commence the conversation around health and wellbeing as an initial entry point.

Distressed men returning high ratings for robot likability and comfort to discuss non-health topics is an interesting finding and suggests that a robot delivering information or advice may be more engaging and acceptable for men than other traditional approaches to engaging men in conversations about mental health. It is known that men are more likely to engage in behaviors that increase health risk, disease and injury even when many of those are preventable. There is a lower rate of health service attendance, meaning that there are fewer opportunities for health education, assessment or intervention from a health professional, placing men in a higher risk category for developing health-related problems [91]. Men are less likely to attend general practitioner appointments or regular health check-up visits, and more likely to delay health service visits during their condition [92-95]. In response, there has been a call to create services that increase the uptake of health information using a friendlier and more convenient format for men [e.g. 93-95, 96]. The use of a social robot to deliver some of those services may be a viable delivery option given its more transactional nature if it meets their preferential need in terms of chosen modality for health-related support services.

## 5.3 Strengths and Limitations of the Trial

A strength for this trial was a large convenience sample to capture first-impression evaluations from a group of individuals who had often not engaged with a social robot before. This led to a broad snapshot of public opinions around a social robot in a health-related role, particularly for those who were frequently accessing a higher education setting. This sample included an even split of genders, a diversity of study degrees, and a broad age range for individuals who were sampled around campus to better understand different robot opinions. A limitation was that most individuals randomized to the wellbeing condition had already undertaken some form of mindfulness practice before, which is a plausible scenario given the rising deployment of mindfulness in high-school based programs [e.g. 97]. Therefore, they may not have felt the need to continue interacting with a robot for this purpose after the session. Future iterations of the program could provide more sophisticated or advanced technique rehearsal for those who are already familiar with the concept, although simple rehearsal and technique refreshers for those who have not practiced in some time might still provide some benefit in itself. In addition, they may not have seen additional benefit or value in completing a guided mindfulness session with the robot if they were already aware of how to use the technique. For those individuals, the process of re-learning the mindfulness technique delivered by the robot may not have been enough or sufficient to improve those scores. For instance, fewer individuals seeing the need to practice mindfulness again using the robot as a guide as reported by their scores. This trial did not involve any follow-up measures beyond the single session and the initial intention of the trial involved an evaluation of receptiveness to participate in a brief robot-delivered session. Requesting individuals to provide personal details for a follow-up may have deterred their initial sign-up, but might instead be presented as an optional addition to the future trial given high participation rates. It should be noted as a limitation that the novelty of the robot alone may have been conducive to high scores, irrespective of the session content. To overcome this challenge, two conditions were implemented. Given that the information or technique rehearsal conditions were similar, it is possible that they were rated because of the robot itself. It is also possible that participants were not aware of other technical use cases that the robot could provide for them, such as the level of interactivity in the control condition. Receptivity to participate in a brief wellbeing technique did appear to be closely tied to their evaluation of the robot that is delivering the session. This signifies that high importance should be placed on robot design, engagement and user experience when delivering an interaction around wellbeing to people, and that the robot is not simply interpreted as a non-influential apparatus to deliver content.

## 6.  Conclusions and Future Work

This pilot randomized controlled trial explored the use of a humanoid social robot to deliver a brief wellbeing technique to a large sample based on a higher education campus. Overall, findings demonstrated initial feasibility and prospective use for a social robot in a healthcare service for wellbeing promotion, which was met with enjoyment, interest and uptake rates. Neither condition outperformed the other. This shows that an initial meeting with a social robot can involve either technique training that is commenced straight away or start with an initial discussion to then later book someone in. This trial marks a stepping stone towards the design and deployment of a high-powered brief robot-delivered mindfulness training program (e.g. treatment-focused randomized controlled trial design). This includes an investigation into the longitudinal effect on individual wellbeing and its related cost-effectiveness to run on a higher education campus. This trial framework also creates the opportunity to build in additional modules that can later be adapted to address higher intensity and multifaceted topics often experienced by university students, such as stress, loneliness, anxious thoughts, or persistent low mood [9, 10].

## Acknowledgements

We would like to thank the research assistants involved in this project for helping with participant recruitment. We would like to thank Belinda Ward and Gary Rasmussen for the provision of testing room space and use of a social robot for the duration of the trial. This research was supported by the Australian Research Council project number CE140100016. Informed consent was obtained from all individual participants included in the study. The authors have no competing interests to declare that are relevant to the content of this article. The data analysed during this study are not publicly available, but de-identified data may be made available on reasonable request.